\pdfoutput=1

\documentclass[11pt]{article}

\usepackage[]{ACL2023}

\usepackage{times}
\usepackage{latexsym}

\usepackage[T1]{fontenc}

\usepackage[utf8]{inputenc}

\usepackage{microtype}

\usepackage{inconsolata}
\usepackage{microtype}
\usepackage{graphicx}
\usepackage{amsmath} 
\usepackage{amsfonts}
\usepackage{booktabs}
\usepackage{multirow}
\usepackage{multicol}
\usepackage{subfig}
\usepackage[ruled,linesnumbered,boxed]{algorithm2e}
\usepackage{hyperref}
\hypersetup{
hidelinks,
colorlinks=true,
allcolors=black,
pdfstartview=Fit,
breaklinks=true
}
\newcommand{\tabincell}[2]{\begin{tabular}{@{}#1@{}}#2\end{tabular}}
\usepackage{bbding}

\title{Preserving Commonsense Knowledge from Pre-trained \\ Language Models via Causal Inference}

\author{Junhao Zheng, Qianli Ma*, Shengjie Qiu, Yue Wu, Peitian Ma, \\ {\bf Junlong Liu, Huawen Feng, Xichen Shang\and Haibin Chen} \\
  School of Computer Science and Engineering, \\
  South China University of Technology, Guangzhou, China\\
  \texttt{junhaozheng47@outlook.com}, 
  \texttt{qianlima@scut.edu.cn}\thanks{*Corresponding author}}

\begin{document}
\maketitle
\begin{abstract}
Fine-tuning has been proven to be a simple and effective technique to transfer the learned knowledge of Pre-trained Language Models (PLMs) to downstream tasks.
However, vanilla fine-tuning easily overfits the target data and degrades the generalization ability.
Most existing studies attribute it to catastrophic forgetting, and they retain the pre-trained knowledge indiscriminately without identifying what knowledge is transferable.
Motivated by this, we frame fine-tuning into a causal graph and discover that the crux of catastrophic forgetting lies in the missing causal effects from the pre-trained data. 
Based on the causal view, we propose a unified objective for fine-tuning to retrieve the causality back.
Intriguingly, the unified objective can be seen as the sum of the vanilla fine-tuning objective, which learns new knowledge from target data, and the causal objective, which preserves old knowledge from PLMs.
Therefore, our method is flexible and can mitigate negative transfer while preserving knowledge.
Since endowing models with commonsense is a long-standing challenge, we implement our method on commonsense QA with a proposed heuristic estimation to verify its effectiveness.
In the experiments, our method outperforms state-of-the-art fine-tuning methods on all six commonsense QA datasets and can be implemented as a plug-in module to inflate the performance of existing QA models. \footnote{Our codes are publicly available at \href{https://github.com/zzz47zzz/CET}{https://github.com/zzz47zzz/CET} and \href{https://github.com/qianlima-lab/CET}{https://github.com/qianlima-lab/CET}} 
\end{abstract}

\section{Introduction}

Deep Pre-trained Language Models (PLMs) such as RoBERTa \citep{liu2019roberta} and T5 \citep{raffel2020exploring}) are inherently knowledge bases since they are exposed to a tremendous amount of data (e.g., the C4 dataset \citep{raffel2020exploring}) in the pre-training stage \citep{petroni-etal-2019-language,alkhamissi2022review}.
Unfortunately, transferring the intrinsic knowledge in PLMs to downstream tasks is non-trivial.
In practice, fine-tuning is adopted widely due to its flexibility \citep{chen-etal-2020-recall} and numerous improved methods \citep{lee2019mixout,chen-etal-2020-recall,chen2019catastrophic,mosbach2020stability,zhang2020revisiting,xu2021raise,aghajanyan2020better,wu-etal-2022-noisytune} are proposed in recent years.
However, fine-tuning faces two challenges when adapting models to new domains \citep{chen2019catastrophic}, including catastrophic forgetting \citep{kirkpatrick2017overcoming} and negative transfer \citep{torrey2010transfer}.
More specifically, catastrophic forgetting refers to models losing previously learned knowledge and overfitting the target domain data. 
Negative transfer occurs because not all pre-trained knowledge is transferable across domains.
Obviously, catastrophic forgetting and negative transfer constitute a dilemma where the crux lies in identifying and utilizing transferable knowledge. 

A large body of previous work has been conducted to solve this problem.
Existing fine-tuning methods for mitigating catastrophic forgetting can be summarized as preventing the fine-tuned models from deviating too far from the pre-trained weights.
For example, \textit{RecAdam} \citep{chen-etal-2020-recall} and \textit{Child-Tuning} \citep{xu2021raise} utilize the Fisher Information Matrix estimated by the pre-trained model to constraint the update in the fine-tuned model.
\textit{Mixout} \citep{lee2019mixout} randomly replaces the model parameters with their pre-trained weights.
These methods constrain the update of models' parameters indiscriminately without identifying what knowledge is transferable and thus susceptible to negative transfer.
\citet{chen2019catastrophic} proposed \textit{BSS}, which focuses on mitigating negative transfer by penalizing the small singular values of the feature matrix.
However, when only negative transfer is concerned, \textit{BSS} may not fully utilize the pre-trained knowledge.

In this paper, we propose a novel method called \textit{\textbf{C}ausal \textbf{E}ffect \textbf{T}uning} (\textbf{CET}) for mining the pre-trained knowledge in PLMs.
Unlike the previous fine-tuning method, our method is rooted in the theory of causal inference. 
It delves into the causalities between data, models, and features instead of merely statistical association.
First, we frame vanilla fine-tuning into a causal graph \citep{glymour2016causal} and find out that the cause of catastrophic forgetting is the vanishing causal effects of pre-trained data.
Therefore, preventing forgetting is to maximize the causal effect.
Then, we approximate the causal effect with the likelihood of the joint prediction of K-Nearest-Neighbor (KNN) samples.
Since equipping models with commonsense knowledge is still challenging, we implement the proposed causal graph with a heuristic approximation on commonsense QA.
We measure the distance with the similarity between gold answers (i.e., ground-truth answers) instead of questions for retrieving KNNs.
The rationale is that the questions with the same gold answer share the same commonsense knowledge in PLMs. 
Finally, we apply our method to RoBERTa \citep{liu2019roberta} and T5 \citep{raffel2020exploring} and conduct extensive experiments on six commonsense datasets.
The experimental results show that our method outperforms state-of-the-art fine-tuning methods and can be plugged into the state-of-the-art QA models to improve performance.

More importantly, our method is lightweight and flexible since it requires no learnable parameter except PLMs and has fewer hyper-parameters to tune. 
It is worth noting that our method readily controls the strength of knowledge preservation by a single hyper-parameter, enabling a good balance between preserving pre-trained knowledge and absorbing new knowledge from downstream tasks.
In summary, our contributions are three-fold:
\begin{itemize}
    \item We present a causal graph for fine-tuning with less forgetting by identifying the root cause of catastrophic forgetting as the missing causal effects of pre-trained data.
    \item Based on the proposed causal graph, we design a lightweight and flexible fine-tuning method called \textit{\textbf{C}ausal \textbf{E}ffect \textbf{T}uning} for preserving knowledge in PLMs. 
    \item For commonsense QA, we estimate the causal effect with a heuristic approximation. And we verify the effectiveness and versatility of our method through extensive experiments on six commonsense QA datasets.
\end{itemize}

\begin{figure*}[!t]
    \centering
    \subfloat[Vanilla Fine-Tuning]{
        \includegraphics[width=0.3\linewidth]{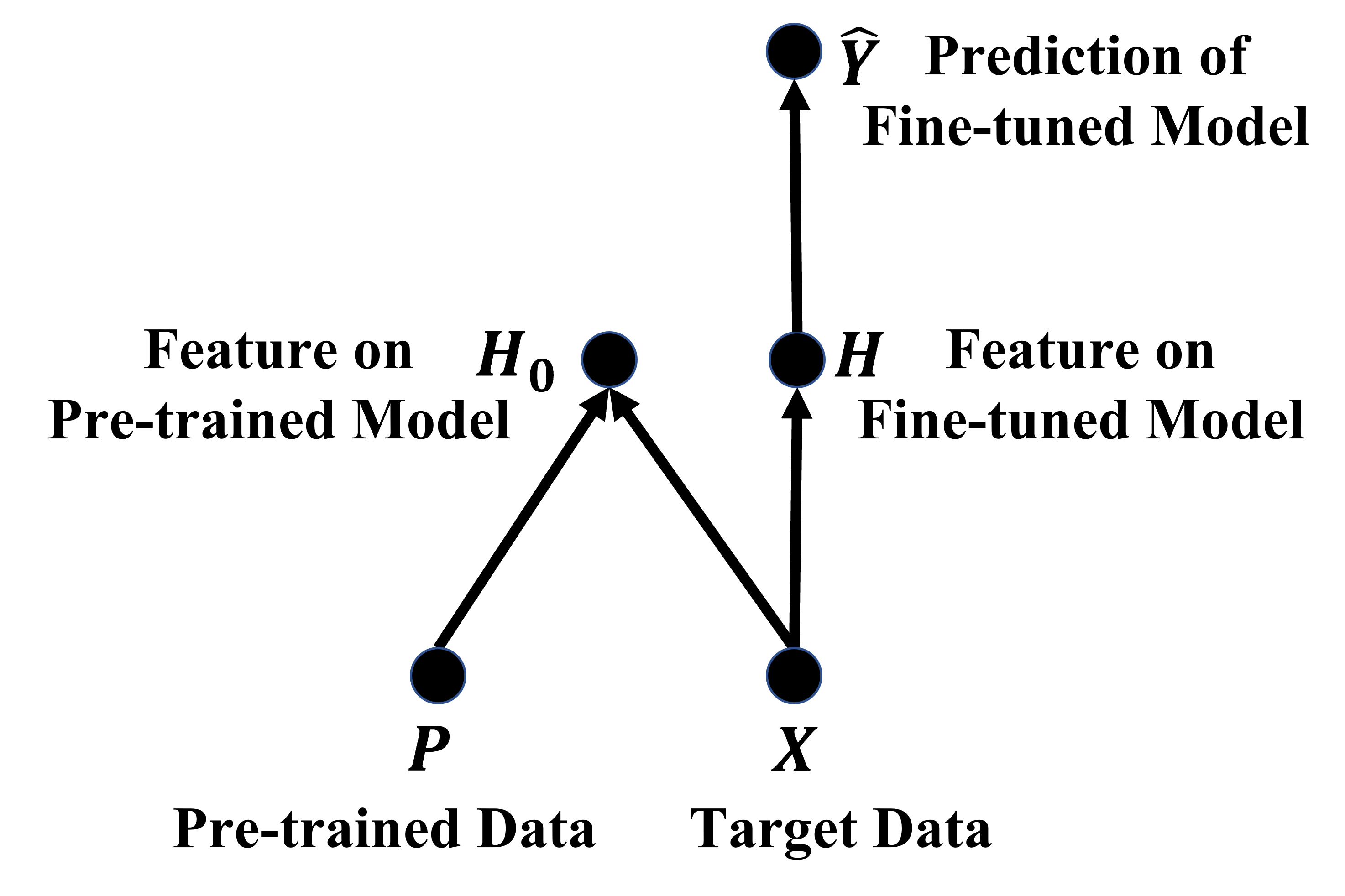}
        \label{fig1a}
    }
    \subfloat[Fine-Tuning with Less Forgetting]{
        \includegraphics[width=0.3\linewidth]{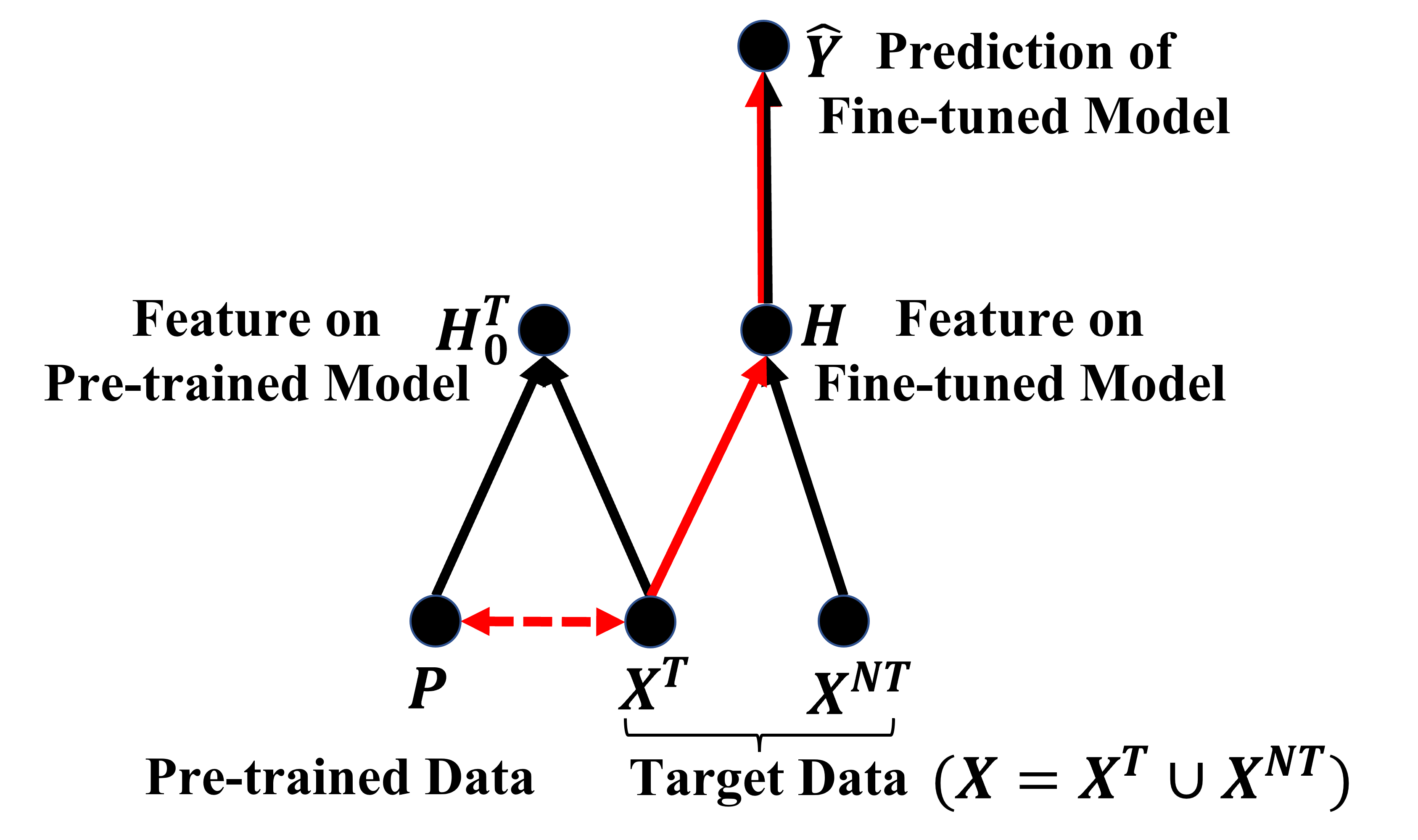}
        \label{fig1b}
    }
    \hspace{0.5cm}
    \subfloat{
        \includegraphics[width=0.2\linewidth]{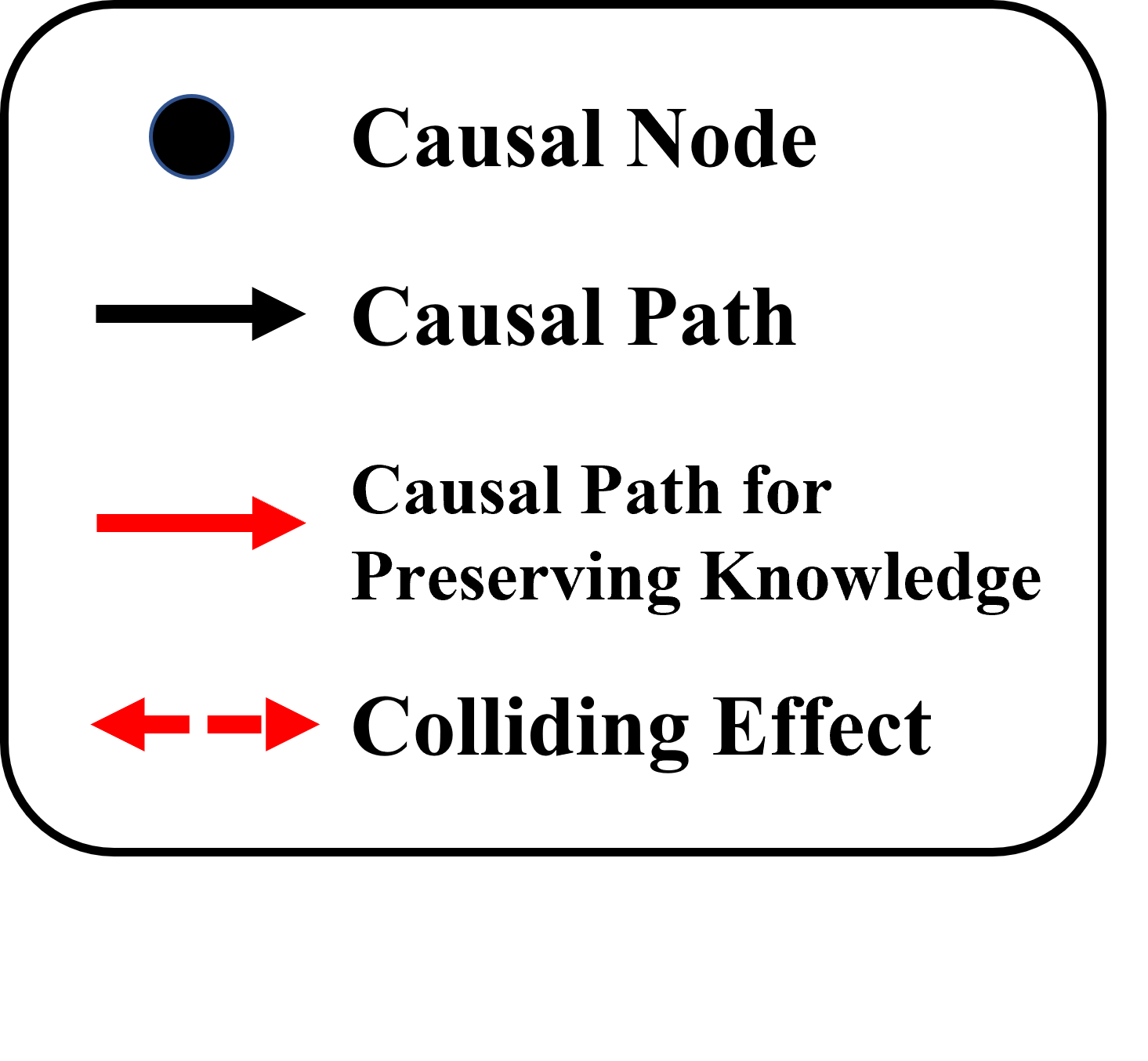}
    }
    \caption{The causal graphs of vanilla fine-tuning and our method. (a): The knowledge is forgotten during vanilla fine-tuning since the causal effect of the pre-trained data is missing; (b): When conditioning on $H_0^T$, the causal effect of the pre-trained data is retained through the causal path $P \leftrightarrow X^T \rightarrow H \rightarrow \hat{Y}$. In addition, the model absorbs new knowledge from $X^{NT}$ through the causal path $X^{NT} \rightarrow H \rightarrow \hat{Y}$.}
\end{figure*}

\section{Related Work}
\subsection{Fine-tuning Methods}
Apart from the methods mentioned above, some approaches improve downstream performances from the perspective of robustness. 
\citet{aghajanyan2020better} proposed \textit{R3F}, which regularizes the symmetric KL divergence between the classifications of the original samples and the perturbed ones.
\citet{wu-etal-2022-noisytune} proposed \textit{Noisytune}, which adds uniform distribution noise to pre-trained parameters before fine-tuning to reduce the risk of overfitting the pre-training tasks and data.
Besides, \citet{mosbach2020stability,zhang2020revisiting} increased the stability of fine-tuning BERT \citep{devlin-etal-2019-bert} in the low-data regime.
\citet{mosbach2020stability} advocated fine-tuning for a long time and choosing good optimizers and hyper-parameters.
\citet{zhang2020revisiting} verified that re-initialized the top layers of BERT helps pre-trained knowledge transfer to downstream tasks.

\subsection{Causal Inference}
Causal inference \citep{glymour2016causal,scholkopf2022causality} has been recently introduced to various computer vision tasks such as image classification \citep{hu2021distilling}, semantic segmentation \citep{zhang2020causal} and long-tailed classification \citep{tang2020long,nan2021uncovering}, and NLP tasks such as distantly supervised NER \citep{zhang2021biasing}, neural dialogue generation \citep{zhu2020counterfactual} and continual named entity recognition \citep{zheng2022distilling}.
To our best knowledge, we are the first to apply causal inference to fine-tuning.

\subsection{Continual Learning}
Although catastrophic forgetting happens in both continual learning \citep{rebuffi2017icarl,hu2021distilling} and fine-tuning, the targets of these two tasks are fundamentally different. 
Continual learning aims to learn a growing number of tasks sequentially and maximize the performance on all recognized tasks.
In contrast, fine-tuning maximize only the performance of target tasks.
The recent advance in continual learning \citep{hu2021distilling,zheng2022distilling} partially inspires this work.

\section{Methodology}

In this section, we first use causal graphs \citep{pearl2009causality} to analyze how pre-trained knowledge is forgotten in fine-tuning.
Then, we present a causal graph for anti-forgetting based on previous analysis.
Next, we estimate the causal effect through derivations and propose a unified learning objective for fine-tuning with less forgetting.
At last, we provide a heuristic approximation for estimating the causal effect on a challenging downstream task, commonsense QA.
Note that the proposed causal graph and the fine-tuning method are generic to all downstream tasks.

\subsection{Vanilla Fine-Tuning}
In a causal graph, nodes represent variables, and directed edges are causalities between nodes.
Fig.(\ref{fig1a}) delineates the process of vanilla fine-tuning.
We denote the pre-trained data (i.e., pre-trained knowledge) as $P$; the data in target tasks as $X$; the feature of $X$ extracted by the pre-trained model and fine-tuned model as $H_0$ and $H$, respectively; the prediction of the fine-tuned model on target tasks as $\hat{Y}$ (i.e., the probability over categories).
The causality between nodes (i.e., directed edges) is as follows:
(1) $X \rightarrow H \rightarrow \hat{Y}$: $X \rightarrow H$ represents that the feature $H$ is extracted by the backbone model such as RoBERTa, and $H \rightarrow \hat{Y}$ represents a classifier compute the prediction $\hat{Y}$ according to the extracted feature $H$;
(2) $X \rightarrow H_0 \leftarrow P$: $H_0$ is determined by both $P$ and $X$ because $H_0$ is extracted by the pre-trained model, which is trained on $P$ \footnote{Here, we ignore the effect of initial parameters initialized from the pre-trained model since it will be exponentially decayed towards zero during fine-tuning \citep{kirkpatrick2017overcoming}.}.

Then, the effect of pre-trained data $P$ on predictions $\hat{Y}$ can be calculated as:
\begin{align}
    \textit{Effect}_{P} \nonumber& = \mathbb{P}(\hat{Y}=\hat{y}|\textit{do}(P=p)) \\
   &- \mathbb{P}(\hat{Y}=\hat{y}|\textit{do}(P=0)) \label{eq1} \\
   &= \mathbb{P}(\hat{Y}=\hat{y}|P=p) - \mathbb{P}(\hat{Y}=\hat{y}|P=0) \label{eq2} \\
   &= \mathbb{P}(\hat{Y}=\hat{y}) - \mathbb{P}(\hat{Y}=\hat{y}) \label{eq3} \\
   &= 0 \label{eq4},
\end{align}
In Eq.(\ref{eq1}), $\textit{do}(P=0)$ represents that no pre-trained data is used for pre-training, and $\textit{do}(P=p)$ represents a standard pre-training is performed.
Then, $\mathbb{P}(\hat{Y}=\hat{y}|\textit{do}(P=p))$ is the prediction given by a \textbf{pre-trained-then-fine-tuned} model and $\mathbb{P}(\hat{Y}=\hat{y}|\textit{do}(P=0))$ is the prediction given by a \textbf{randomly-initialized-then-fine-tuned} model.
Eq.(\ref{eq1}) defines $\textit{Effect}_{P}$ as the difference between the two predictions.
Eq.(\ref{eq2}) holds because $P$ has no parent nodes.
Eq.(\ref{eq3}) holds because collider $H_0$ blocks all causal paths from $P$ to $Y$.

Eq.(\ref{eq1})-(\ref{eq4}) shows that a vanilla fine-tuned model will eventually forget all pre-trained knowledge when no constraints are imposed. 
In practice, fine-tuned models will not forget all learned knowledge because the learning rate and training time are considerably lower and shorter than those in pre-training.
However, fine-tuned models likely forget partial pre-trained knowledge, overfit the target data, and fall into sub-optimal states since the amount of target data is usually considerably less than that of pre-trained data.

\subsection{Fine-Tuning with Less Forgetting}
The causal graph in Fig.(\ref{fig1a}) necessitates the retrieval of the causality between $P$ and $\hat{Y}$ back.
A straightforward solution is utilizing the pre-trained data to constrain model behaviors in new tasks.
However, it is often obstructed by time, space, and financial constraints.

Thanks to causal inference, we can build a causal path between $P$ and $X$ without storing $P$.
In the causal graph Fig.(\ref{fig1a}), $H_0$ is the joint outcome of the independent causes $P$ and $X$.
Intriguingly, once the common effect $H_0$ is observed, the causes $P$ and $X$ become dependent.
The causal effect is called \textbf{colliding effect} in \citet{hu2021distilling,zheng2022distilling}\footnote{This phenomenon is also known as \textit{Berkson's paradox} in \citep{berkson1946limitations} and as the \textit{explaining away effect} in \citep{peari1983computationai}.}.
We'd like to provide a vivid example \citep{pearl2009causality} for understanding this pattern in causal inference:
If the admission criteria to a certain school require either high grades or special musical talents, then these two attributes will be found to be correlated (negatively) in that school's student population, even if these attributes are uncorrelated in the population at large.
By conditioning on $H_0$, the causal effect of pre-trained data is preserved during fine-tuning (i.e., $ \textit{Effect}_{P}>0$), and thus the pre-trained knowledge is preserved.

Except for preserving old knowledge, assimilating new knowledge from target data is critical.
In addition, negative transfer may occur if we preserve pre-trained knowledge overly.
Motivated by this, we split the target data into two nodes $X^{T}$ and $X^{NT}$.
$X^{T}$ represents the samples where we calculate colliding effects, and their knowledge should be transferred from PLMs.
$X^{NT}$ is the samples where we do not calculate colliding effects, and their knowledge is domain-specific and should be absorbed into fine-tuned models.
Consequently, the causal graph for our method is in Fig.(\ref{fig1b}), and the rationale is as follows:
The fine-tuned model preserves pre-trained knowledge by utilizing colliding effects ($P \leftrightarrow X^{T}$) while learning domain-specific knowledge ($X^{NT}$).
The final prediction depends on both \textbf{pre-trained knowledge} and \textbf{domain-specific knowledge} from causal paths $P \leftrightarrow X^{T} \rightarrow H  \rightarrow{\hat{Y}}$ and $X^{NT} \rightarrow H \rightarrow \hat{Y}$, respectively.

\begin{figure*}[t]
    \centering
    \includegraphics[width=0.75\linewidth]{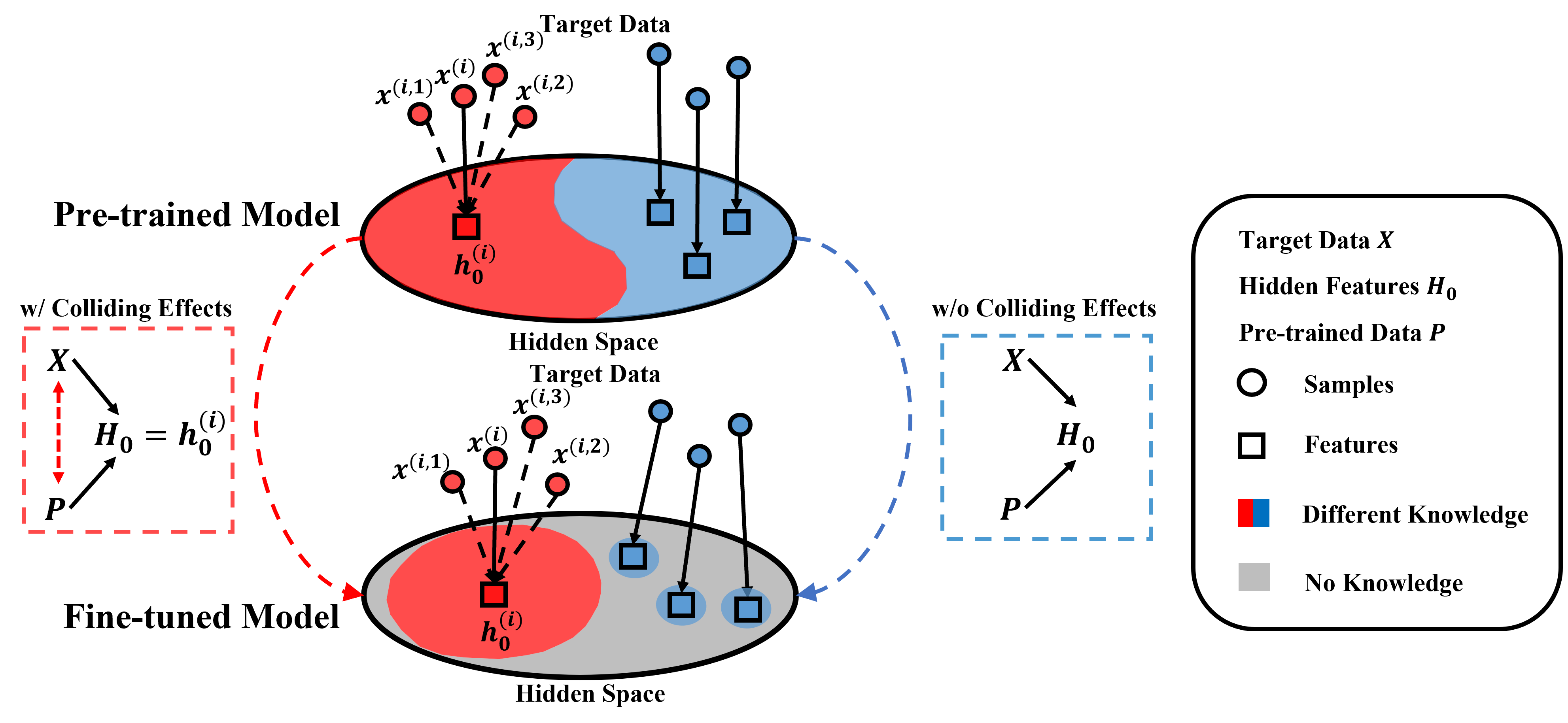}
    \caption{An illustration of Causal Effect Tuning. $x^{(i)}$ is the anchor sample and $h_0^{(i)}$ is the hidden feature extracted by the pre-trained model. $x^{(i,1)},x^{(i,2)},x^{(i,3)}$ are the KNNs of $x^{(i)}$. We apply colliding effects on $x^{(i)}$ to preserve the old knowledge. After fine-tuning, the ``{\color{red}Red}'' knowledge is preserved with colliding effects, and ``{\color{blue}blue}'' knowledge is forgotten without colliding effects. A specific instance is as follows: $x^{(i)}$= ``What is a fast but expensive way to send small cargo? (answer: airplane)''; $x^{(i,1)}$=``Where could you find a seat that sometimes vibrates?''(answer: airplane); $x^{(i,2)}$=``What has metal wings?''(answer: airplane); $x^{(i,3)}$= "It was important precious cargo, so it was delivered as quickly as possible by means of what?''(answer: aeroplane). The ``{\color{red}red}'' knowledge represents the commonsense about ``\textbf{airplane}''.}
    \label{fig2}
\end{figure*}

\subsection{Estimating Colliding Effects}
Next, we need to estimate the colliding effect between $P$ and $X^{T}$.
When conditioning on $H_0$, $\textit{Effect}_{P}$ can be calculated as:
\begin{align}
   &\textit{Effect}_{P} = \sum_{i=1}^N \textit{Effect}_{P}^{(i)} \label{eq5} \\
   &\approx \sum_{i=1}^N \sum_{k=0}^K \mathbb{P}(\hat{Y}^{(i)}|X=x^{(i,k)}) W_P(x^{(i)},x^{(i,k)}), \label{eq6}
\end{align}
where $\sum_{k=0}^K W_P(x^{(i)},x^{(i,k)}) = 1$.
$N$ is the number of samples in the target data and $x^{(i)}$ is the $i$-th sample.
$\textit{Effect}_{P}^{(i)}$ is the colliding effect of $P$ on the prediction $\hat{Y}^{(i)}$.
$W_P(\cdot, \cdot)$ is a function determined by the pre-trained model and measures the similarity between two samples in the hidden space of the pre-trained model.
In this case, we denote $W_P(x^{(i)},x^{(i,k)})$ as $W_{i,k}$ for brevity.
$x^{(i,k)}$ is the $k$-th nearest neighbor of $x^{(i)}$ in the hidden space.
Since $x^{(i)}$ always has the largest similarity with itself, we let $x^{(i,0)}=x^{(i)}$ and call $x^{(i)}$ the anchor sample.
Besides, we assume that the $K$ Nearest Neighbours (KNNs) are sorted in descending order according to the similarity.
Therefore, we have $W_{i,0} \ge W_{i,1}\ge W_{i,2} \ge \cdots \ge W_{i,K}$.
$K$ is a hyper-parameter representing the number of neighbors for estimating $\hat{Y}^{(i)}$.
We provide a detailed derivation and further explanation in Appendix \ref{sec:appendix_a}.

Eq.(\ref{eq5}) re-writes the total causal effect as the sum of the causal effect on the prediction of each target sample (i.e.,$\textit{Effect}_{P}^{(i)}$).
In Eq.(\ref{eq6}), $\mathbb{P}(\hat{Y}^{(i)}|X=x^{(i,k)})$ represents the likelihood of $\hat{Y}^{(i)}$ when $x^{(i,k)}$ is the model input.
Eq.(\ref{eq6}) shows that $\textit{Effect}_{P}^{(i)}$ can be approximated by the weighted sum of the likelihood when the model input is the anchor sample $x^{(i)}$ and its KNNs.
Since we expect to maximize $\mathbb{P}(\hat{Y}^{(i)}=y^{(i)}|X=x^{(i)})$, maximizing $\textit{Effect}_{P}^{(i)}$ equals to maximizing the likelihood of the \textbf{joint prediction} on the ground-truth label $y^{(i)}$.

\subsection{Overall Objective}
In Eq. \ref{eq6}, the total causal effect $\textit{Effect}_P$ is broken down into the causal effect of each sample $\textit{Effect}_{P}^{(i)}$.
In this case, maximizing $\textit{Effect}_P$ is to preserve the related knowledge of all samples.
As we mentioned before, indiscriminately preserving knowledge may lead to negative transfer.
To address this problem, we introduce a similarity threshold $\theta$ to select the number of nearest neighbors for each sample automatically.
Specifically, for the $i$-th sample, we truncate the $k_i$ ($K \ge k_i \ge 0$) nearest neighbors whose similarity is greater or equal than $\theta$.
In this way, we differentiate the strength of knowledge preservation on each sample by selecting the neighbors with small distances to their anchor sample.
More interestingly, when $k_i=0$, \textit{i.e.}, a sample has no neighbors, the $\textit{Effect}_{P}^{(i)}$ amounts to $\mathbb{P}(\hat{Y}^{(i)}=y^{(i)}|X=x^{(i)})$, which is exactly the objective of each sample in vanilla fine-tuning.
Fig. \ref{fig2} provides an illustration for our method, where the samples with no neighbors can be seen as a special case of our method.
Formally, we define the overall objective as follows:
\begin{align}
   & \text{max} \quad \textit{Effect}_P = \sum_{i=1}^N \textit{Effect}_{P}^{(i)} 
\label{eq7} \\
   &= \underbrace{ \sum_{i \in \mathcal{S}^{T}} \textit{Effect}_{P}^{(i)} }_{\text{Colliding Effects}}
   + \underbrace{ \sum_{i \in \mathcal{S}^{NT}} \textit{Effect}_{P}^{(i)} }_{\text{Vanilla Fine-Tuning}}, \label{eq8} \\
   &= \underbrace{ \sum_{i \in \mathcal{S}^{T}} \sum_{k=0}^{k_i} \mathbb{P}(\hat{Y}^{(i)}|X=x^{(i,k)}) W_{i,k}  }_{\text{Colliding Effects}} \label{eq9} \\
   & \nonumber \qquad + \underbrace{ \sum_{i \in \mathcal{S}^{NT}} \mathbb{P}(\hat{Y}^{(i)}|X=x^{(i)}) }_{\text{Vanilla Fine-Tuning}},
\end{align}
where $\sum_k{W_{i,k}}=1, \mathcal{S}^{T}=\{i|k_i>0\}, \mathcal{S}^{NT}=\{i|k_i=0\}$.
Considering the distances between KNNs and their anchor sample are approximated and thus inaccurate, we set $W_{i,0}=W_0$ and $W_{i,1}=W_{i,2}=\cdots=W_{i,k_i}=\frac{1-W_0}{k_i}$ when $k_i > 0$ for implementation.
$W_0$ is a hyper-parameter for controlling the strength of colliding effects.
When $W_0=0$, the overall target degenerates to the vanilla fine-tuning target.
When $W_0=1$, the overall target retains knowledge indiscriminately on all samples.
In Eq.(\ref{eq9}), the second term amounts to the vanilla fine-tuning objective since only the anchor sample's prediction is computed.
In other words, we preserve knowledge for the samples with KNNs and learn new knowledge for the samples without KNNs.
The rationale is that the knowledge should be preserved when more samples require it to answer the question.
In the proposed causal graph in Fig.(\ref{fig1b}), the first and the second term of Eq.(\ref{eq9}) correspond to the two causal paths through $X^T$ and $X^{NT}$ respectively.
We summarized the proposed method in Fig. \ref{fig2} and Alg. \ref{alg1} in Appendix \ref{sec:appendix_a}. 

\subsection{An Implementation on Commonsense QA}
In this subsection, we provide an implementation for the causal graph in Fig.(\ref{fig1b}) on commonsense QA.
We note that the overall objective in Eq. \ref{eq9} is agnostic to specific downstream tasks and model architectures.
The implementation can be different in various tasks or model architectures, and the key is to find proper KNNs. 
This paper provides an implementation on commonsense QA since PLMs may be endowed with commonsense knowledge in pre-training \citep{petroni-etal-2019-language,alkhamissi2022review}, and it is still challenging for models to capitalize on commonsense \citep{talmor2018commonsenseqa}.

We first formulate the commonsense QA as follows:
Given a dataset with $N$ samples $\{(q^{(i)}, a^{(i)}, \{o_j^{(i)}\}_j)\}_i^N$, we train the best model for choosing the gold answer $a^{(i)}$ among options $\{o_{j}^{(i)}\}$ given a question $q^{(i)}$.
More specifically, the input of the $i$-th sample can be $x^{(i)} = q^{(i)}||o_1^{(i)}||\cdots||o_j^{(i)}$ or $\{x^{(i)}\}_j = \{q^{(i)}||o_j^{(i)}\}_j$ \footnote{Concatenating all options or each option depends on models.} where $||$ is the string-level concatenation.

Then, we define a metric to search KNNs.
A simple solution is to compute the euclidean distance or cosine similarity between the average last hidden states of PLMs.
However, this method struggles to capture accurate semantic meanings, and measuring sentence similarity remains challenging.
In this regard, we provide a simple heuristic approximation.
In most cases, the questions with the same gold answers share the same knowledge.
For example, ``airplane'' is the gold answer to the following questions, and we can use the knowledge about ``airplane'' to answer them:
``\textit{What is a fast but expensive way to send small cargo?}''; ``\textit{Where could you find a seat that sometimes vibrates?}''; ``\textit{What has metal wings?}''.
Therefore, we estimate the similarity between gold answers to cope with the difficulty of evaluating sentence similarity.
Since options are usually much shorter than questions, lightweight tools such as spaCy \citep{Honnibal_spaCy_Industrial-strength_Natural_2020} can be used to retrieve gold answers with close semantic meanings (e.g., ``airplane'' and ``aeroplane'').

At last, we define the input of the $i$-th sample's KNNs as $x^{(i,k)}=q^{(i,k)}||o_1^{(i)}||\cdots||o_j^{(i)}$ or $\{x^{(i,k)}\}_j = \{q^{(i,k)}||o_j^{(i)}\}_j$.
It alleviates the overfitting problem since the model needs to select the correct answer among the options of anchor sample when the question is from its KNNs. 

\begin{table*}[!t]
  \small
  \centering
  \caption{Comparison with state-of-the-art methods. The average accuracy (\%) and the standard derivation are reported.}
  \resizebox{\linewidth}{!}{
    \begin{tabular}{lccccccc}
    \toprule
    Methods & CSQA  & OBQA  & ARC-Easy & ARC-Challenge & QASC  & PIQA  & SIQA \\
    \midrule
    Fine-Tuning & 75.74 (0.47) & 68.12 (0.32) & 67.66 (0.45) & 45.98 (0.53) & 54.04 (1.05) & 78.62 (0.53) & 77.46 (0.33) \\
    BSS   & 76.21 (0.63) & 68.64 (1.23) & 68.24 (0.31) & 46.62 (0.80) & 53.82 (1.20) & 78.20 (0.96) & 77.35 (0.18) \\
    ChildTune-F & 75.50 (0.44) & 69.84 (0.88) & 68.17 (0.77) & 46.30 (1.67) & 54.41 (1.63) & 77.61 (1.06) & 75.87 (0.64) \\
    ChildTune-D & 76.76 (0.81) & 69.36 (0.60) & 67.86 (0.73) & 45.28 (0.67) & 55.77 (0.52) & 78.32 (0.38) & 78.20 (0.35) \\
    Mixout & 76.09 (0.56) & 69.70 (0.71) & 67.85 (0.57) & 44.87 (0.72) & 57.34 (1.02) & 79.22 (0.31) & 77.89 (0.37) \\
    NoisyTune & 76.01 (0.61) & 67.56 (0.52) & 67.61 (0.58) & 46.05 (0.65) & 54.43 (0.60) & 78.61 (0.31) & 76.59 (0.36) \\
    R3F   & 76.59 (0.48) & 68.47 (0.26) & 68.13 (0.68) & 47.01 (0.58) & 55.69 (0.78) & 79.38 (0.60) & 77.05 (0.44) \\
    RecAdam & 75.43 (0.33) & 70.68 (0.89) & 68.07 (0.69) & 45.90 (0.59) & 54.62 (1.22) & 78.26 (1.25) & 76.71 (0.61) \\
    ReInit & 75.51 (0.71) & 69.92 (1.14) & 67.63 (0.59) & 46.68 (0.39) & 52.12 (1.66) & 78.61 (0.37) & 77.79 (0.15) \\
    CET(Ours) & \textbf{76.82 (0.33)} & \textbf{70.76 (0.33)} & \textbf{68.53 (0.53)} & \textbf{47.52 (0.38)} & \textbf{57.57 (0.44)} & \textbf{79.43 (0.27)} & \textbf{78.76 (0.31)} \\
    \bottomrule
    \end{tabular}%
    }
  \label{tab:table1}%
\end{table*}%

\begin{table*}[htbp]
  \small
  \centering
  \caption{Comparisons with knowledge-graph-based methods on CSQA with different proportions of training data. We use the train-dev-test split in \citet{jiang2022great} and thus the CSQA results are inconsistent with those in other experiments. The results of RoBERTa-large, RGCN, KagNet, Relation Network, MHGRN, QAGNN, and SAFE are reported in \citet{jiang2022great}. We report the average accuracy (\%).}
        \begin{tabular}{l|c|c|cccccc}
    \toprule
    \multicolumn{1}{c|}{\multirow{2}[4]{*}{Methods}} & \multirow{2}[4]{*}{use GNN?} & \multirow{2}[4]{*}{use KG?} & \multicolumn{6}{c}{Proportion of Training Data} \\
\cmidrule{4-9}          &       &       & 5\%   & 10\%  & 20\%  & 50\%  & 80\%  & 100\% \\
    \midrule
    RoBERTa-large & \XSolidBrush & \XSolidBrush & 29.66  & 42.84  & 58.47  & 66.13  & 68.47  & 68.69  \\
    \midrule
    +RGCN \citep{schlichtkrull2018modeling} & \Checkmark & \Checkmark & 24.41  & 43.75  & 59.44  & 66.07  & 68.33  & 68.41 \\
    +KagNet \citep{lin-etal-2019-kagnet} & \Checkmark & \Checkmark & 21.92  & 49.83  & 60.09  & 66.93  & 69.14  & 68.59 \\
    +Relation Network \citep{santoro2017simple} & \Checkmark & \Checkmark & 23.77  & 34.09  & 59.90  & 65.62  & 67.37  & 69.08 \\
    +MHGRN \citep{feng-etal-2020-scalable} & \Checkmark & \Checkmark & 29.01  & 32.02  & 50.23  & 68.09  & 70.83  & 71.11 \\
    +QAGNN \citep{yasunaga-etal-2021-qa} & \Checkmark & \Checkmark & 32.95  & 37.77  & 50.15  & 69.33  & 70.99  & 73.41 \\
    +SAFE \citep{jiang2022great} & \Checkmark & \Checkmark & 36.45  & 56.51  & 65.16  & 70.72  & 73.22  & 74.03 \\
    \midrule
    +CET(Ours) & \XSolidBrush & \XSolidBrush & 56.24  & 59.55  & 65.19  & 67.93  & 70.02  & 70.99 \\
    +CET+QAGNN & \Checkmark & \Checkmark & 58.78  & 60.35  & 65.59  & 70.43  & 72.04  & 73.81 \\
    +CET+SAFE & \Checkmark & \Checkmark & \textbf{59.39}  & \textbf{61.02}  & \textbf{65.75}  & \textbf{70.79}  & \textbf{73.31}  & \textbf{74.54} \\
    \bottomrule
    \end{tabular}%
  \label{tab:table7}%
\end{table*}%

\begin{table}[htbp]
  \tiny
  \centering
  \caption{An CSQA example and its KNNs in our method.}
    \begin{tabular}{ccl}
    \toprule
          & Gold Answer & \multicolumn{1}{c}{Question} \\
    \midrule
    Anchor & pet shops & \tabincell{l}{Too many people want exotic snakes. The demand\\ is driving what to carry them?} \\
    \midrule
    \multirow{5}[10]{*}{KNNs} & pet shops & Where can a person buy a snake? \\
\cmidrule{2-3}          & pet shop & Where might  a  blowfish be kept? \\
\cmidrule{2-3}          & pet shop & Where can you take home a hermit crab? \\
\cmidrule{2-3}          & pet store & Where would you get a dog if you do not have one? \\
\cmidrule{2-3}          & pet store & \tabincell{l}{John loves animals and he hates animal abuse.  Because\\ of this, john is very careful about the places he goes. Where\\ might he avoid going?} \\
    \bottomrule
    \end{tabular}%
  \label{tab:table4}%
\end{table}%

\section{Experiments}
\subsection{Settings}
\noindent\textbf{Datasets.}\quad 
We conduct experiments on 6 datasets: CommonsenseQA(CSQA) \citep{talmor2018commonsenseqa}, OpenBookQA(OBQA) \citep{mihaylov-etal-2018-suit}, ARC \citep{clark2018think,clark2016combining}, QASC \citep{khot2020qasc}, SocialIQA (SIQA) \citep{sap-etal-2019-social}, PIQA \citep{bisk2020piqa}.
Since the official test sets of CSQA, QASC, SIQA, and PIQA are not available, we follow \citep{yasunaga-etal-2021-qa} and use the official dev sets as test sets and split in-house dev set from the original training sets.
The dataset statistics are summarized in Table \ref{tab:dataset_statistic} in Appendix \ref{sec:appendix_b}.

\noindent\textbf{Training.}\quad
Given its popularity, we use RoBERTa-large \citep{liu2019roberta} as the backbone model in default.
We also explore T5-large \citep{raffel2020exploring} since \citet{khashabi-etal-2020-unifiedqa} showed that it excels at answering questions in different formats.
Other training details are specified in Appendix \ref{sec:appendix_b}.

\noindent\textbf{Competitive Methods.}\quad
We make comparisons with nine state-of-the-art fine-tuning methods:
vanilla fine-tuning, BSS \citep{chen2019catastrophic}, ChildTune-F\&ChildTune-D \citep{xu2021raise}, Mixout \citep{lee2019mixout}, NoisyTune \citep{wu-etal-2022-noisytune}, R3F \citep{aghajanyan2020better}, RecAdam \citep{chen-etal-2020-recall} and ReInit \citep{zhang2020revisiting}.
For each method, we use the recommended hyper-parameters in the paper and source code for a fair comparison.
We discuss the implementation details of the fine-tuning methods in Appendix \ref{sec:appendix_c}.

\noindent\textbf{Hyper-Parameters.}\quad
As for the hyperparameters of our methods, we fix $K=5$ and search the best $W_0$ in \{0.5, 0.7, 0.9, 0.95, 0.97\} for each dataset.
We use spaCy to estimate the similarity between gold answers.  
We set $\theta=0.99$ for PIQA and $\theta=1.00$ for other datasets (i.e., the gold answers should be matched precisely).

\subsection{Results and Analyses}
\noindent\textbf{Comparisons with State-Of-The-Art.}\quad
To demonstrate the effectiveness of our method, we re-implement several strong baselines on commonsense QA datasets using their officially released codes and hyper-parameters.
The results are summarized in Table \ref{tab:table1}.
Results show that our method outperforms all fine-tuning methods consistently.
On QASC and OBQA, our method achieves 57.57\% and 70.76\% accuracy, obtaining 3.53\% and 2.64\% improvements on vanilla fine-tuning.

\textit{Why our method better preserves commonsense knowledge from PLMs?}
The reasons are two-fold.
The first reason is that our method utilizes the colliding effect for transferring the ``colliding'' commonsense knowledge, while other methods do not.
For instance, in Fig.\ref{fig2}, our method encourages models to update $x^{(i)}$ and its KNNs $x^{(i,1)},x^{(i,2)},x^{(i,3)}$ simultaneously.
In this way, the commonsense knowledge about ``airplane'' that ``airplanes deliver small and precious cargo'', ``airplanes have metal wings'' and ``airplanes have seats'' can be transferred jointly, which reduces the risk of over-fitting.
We provide more examples from each dataset in Table \ref{tab:table4} and Table \ref{tab:table6},\ref{tab:table5}, in Appendix \ref{sec:more_examples_of_colliding_effects}.
The second reason is that our method does not directly constrain (e.g., ChildTune-D, Mixout, RecAdam) or modify (e.g., NoisyTune, ReInit) the parameters of fine-tuned models.
Empirical results show that these methods encounter negative transfers on some of the datasets.
Instead, our method builds upon the causal inference theory and utilizes the joint prediction as a soft constraint to transfer related knowledge while mitigating negative transfer.

\noindent\textbf{Compared with Knowledge-Graph-Based Methods.}\quad
Utilizing knowledge graphs such as ConceptNet \citep{speer2017conceptnet} is a common practice for building commonsense QA systems.
We compared our method with six knowledge-graph-based methods:
Relation Network \citep{santoro2017simple}, KagNet \citep{lin-etal-2019-kagnet}, RGCN\citep{schlichtkrull2018modeling}, MHGRN\citep{feng-etal-2020-scalable}, QAGNN\citep{yasunaga-etal-2021-qa}, SAFE\citep{jiang2022great}.
Detailed descriptions and other related works are given in Appendix \ref{sec:related_works_of_commonsense_QA}.
Note that these methods utilize knowledge graphs (KGs) as external knowledge resources, and most of them train graph neural networks (GNNs) for extracting features from KGs.
In contrast, our method does not introduce any additional learnable parameters except PLMs and the final fully-connected layer. 
The result in Table \ref{tab:table7} shows that our method outperforms RGCN, KagNet, and Relation Network by only mining the internal knowledge of PLMs.
Furthermore, our method significantly outperforms all the knowledge-graph-based methods under low resource conditions ($\leq$ 20\% training data is used), which shows that our method helps PLMs adapt to downstream tasks with less data.

In addition, our method can be easily implemented as a plug-in module by simply substituting the vanilla fine-tuning objective for the causal effect in Eq.(\ref{eq9}).
We combine our method with QAGNN and SAFE, respectively. 
Table \ref{tab:table7} shows that our approach consistently improves QAGNN and SAFE and achieves superior performances.
Therefore, the pre-trained commonsense knowledge benefits downstream tasks even when KGs are introduced.

\noindent\textbf{Fine-tuning on a Cyclic Chain of Tasks.}\quad
To understand how our method preserves knowledge during fine-tuning, we follow \citet{aghajanyan2020better} and design a cyclic chain of tasks:
$$
\underbrace{ A \rightarrow B \rightarrow C}_{Cycle 1}\rightarrow  \underbrace{ A \rightarrow B \rightarrow C}_{Cycle 2}\rightarrow\cdots
$$
In our experiment, we set A=CSQA, B=OBQA, and C=QASC for a demonstration.
Specifically, we start from a PLM and fine-tune it on CSQA.
Then, we use the model fine-tuned on CSQA to initialize the backbone model's parameters and continue fine-tuning it on OBQA.
Table \ref{tab:table9} shows that our method retains knowledge significantly better than vanilla fine-tuning.
The performances on OBQA and QASC improve at every cycle, suggesting that our method effectively retains knowledge from the previous datasets.
Unfortunately, both performances of vanilla fine-tuning and our method on CSQA degrade slightly, showing that negative transfer happens. 
In this case, vanilla fine-tuning will lead to more serious performance degradation.
The experiment is for demonstration, and a better combination of tasks that promote each other may be found.

\noindent\textbf{Ablation Study.}\quad
To verify the effectiveness of our method, we consider the following ablated version of our method:
(1) replacing the KNNs (\textit{Large},Ours) with randomly selected samples (\textit{Rand}) or samples with the smallest similarity (\textit{Small});
(2) searching the KNNs according to the similarity of average last hidden states (\textit{Avg}) instead of gold answers (\textit{Gold}, Ours).
The result in Table \ref{tab:table2} shows that the model learns commonsense knowledge better when the KNNs share the gold answer with close meaning.

\noindent\textbf{Additional Experiments.}\quad
Due to space constraints, we present the experiments on T5, the hyper-parameter analysis, the experiments on Named Entity Recognition, and further discussions in Appendix \ref{sec:additional_experimental_results}.

\begin{table}[t]
  \small
  \centering
  \caption{The results of cyclical sequential fine-tuning for three cycles. The average accuracy (\%) is reported.}
    \begin{tabular}{c|l|cc}
    \toprule
          & Dataset & Fine-Tuning & CET(Ours) \\
    \midrule
    \multirow{3}[2]{*}{Cycle1} & CSQA  & 75.74  & \textbf{76.82 } \\
          & OBQA  & 68.80  & \textbf{70.89 } \\
          & QASC  & 54.31  & \textbf{57.49 } \\
    \midrule
    \multirow{3}[2]{*}{Cycle 2} & CSQA  & 75.52  & \textbf{76.69 } \\
          & OBQA  & 69.95  & \textbf{71.18 } \\
          & QASC  & 55.06  & \textbf{57.64 } \\
    \midrule
    \multirow{3}[2]{*}{Cycle 3} & CSQA  & 75.44  & \textbf{76.75 } \\
          & OBQA  & 70.28  & \textbf{71.45 } \\
          & QASC  & 55.12  & \textbf{57.78 } \\
    \bottomrule
    \end{tabular}%
  \label{tab:table9}%
\end{table}%

\begin{table}[!t]
  \small
  \centering
  \caption{The ablation study of our method. \textit{Gold}/\textit{Avg}: searching the KNNs according to the similarity of gold answers or the average last hidden states. \textit{Large}/\textit{Small}/\textit{Rand}: searching the KNNs with the largest or smallest similarity, or randomly. The average accuracy (\%) is reported.}
    \begin{tabular}{lccc}
    \toprule
    \multicolumn{1}{c}{Methods} & CSQA  & OBQA  & QASC \\
    \midrule
    Gold+Large(Ours) & \textbf{76.82 } & \textbf{70.76 } & \textbf{57.57} \\
    \midrule
    Gold+Rand & 74.61  & 68.53  & 55.77  \\
    Gold+Small & 74.04  & 64.67  & 53.13  \\
    Avg+Large & 76.17  & 69.64  & 55.62  \\
    Avg+Rand & 74.12  & 68.54  & 54.54  \\
    Avg+Small & 74.20  & 68.07  & 53.46  \\
    \midrule
    Fine-Tuning & 75.74  & 68.12  & 54.04  \\
    \bottomrule
    \end{tabular}%
  \label{tab:table2}%
\end{table}%

\section{Conclusion}
We propose a novel fine-tuning technique rooted in causal inference for preserving pre-trained knowledge from PLMs.
Although many fine-tuning methods have been proposed in recent years, most of them overlooked one or both hidden issues of fine-tuning, catastrophic forgetting and negative transfer, which result in a dilemma.
In this paper, we provide an answer to the dilemma from the casual lens.
Impressively, we empirically find that the proposed method achieves the best performance on six commonsense QA datasets and is flexible to be applied to various QA systems and model architectures.

\section*{Limitations}
There are three limitations on our method. 
First, we did not verify our method on more generic tasks, such as text classification, yet it is not limited to commonsense QA.
Extending our method to other downstream tasks is our future work.
Second, our method requires a longer training time and a larger GPU memory since the KNNs require forward and backward propagation additionally.
Third, we do not consider the ambiguity of gold answers, which may affect the quality of KNNs.
For example, ``apple'' may refer to a kind of fruit or a technology company.


\section*{Acknowledgements}
The work described in this paper was partially funded by the National Natural Science Foundation of China (Grant Nos. 62272173, 61872148), the Natural Science Foundation of Guangdong Province (Grant Nos. 2022A1515010179, 2019A1515010768).

\bibliography{ref}
\bibliographystyle{acl_natbib}

\appendix

\section{A Detailed Derivation for the Colliding Effect}
\label{sec:appendix_a}

\SetKwInOut{KwResult}{Require}

\begin{algorithm}[htbp] 
\caption{Causal Effect Tuning} \label{alg1}
\KwIn{
$\mathcal{D}=\{(x^{(i)},y^{(i)})\}_{i=1}^{N}$: a training set with $N$ samples; 
$\mathcal{F}_0$: a pre-trained model
}
\KwOut{
$\mathcal{F}$: a fine-tuned model
}
Initialize $\mathcal{F} \leftarrow \mathcal{F}_0$\;
Compute the KNNs for each sample $x^{(i)}$: $x^{(i,1)},\cdots,x^{(i,k_i)}$\;  

\While{not converge}
{
    Compute $\textit{Effect}_{P}$ according to Eq.(\ref{eq9})\;
    $\mathcal{F} \leftarrow \mathop{\arg\max}\limits_{\mathcal{F}}\textit{Effect}_{P}$\;
}
   
\Return $\mathcal{F}$\;
\end{algorithm}

Without loss of generality, we first define the fine-tuning process formally as follows:
Given a pre-trained model $\mathcal{F}_0$ and a dataset with $N$ samples $\{(x^{(i)}, y^{(i)}\}_{i=1}^N$, we aim to learn a model $\mathcal{F}$ which has the best performance on predicting the label $y^{(i)}$.
Recall that in Eq.(\ref{eq5}), we re-write $\textit{Effect}_{P}$ as the sum of the causal effect on each prediction $\hat{Y}^{(i)}$.
Now, the outcome node $\hat{Y}$ in the causal graph becomes $\hat{Y}^{(i)}$.
Then, we need to condition on $H_0$ to utilize colliding effects.
Considering when predicting $\hat{Y}^{(i)}$, $x^{(i)}$ should play an important role.
Furthermore, when $X=x^{(i)}$, its hidden feature is simply calculated as $h_0^{(i)}=\mathcal{F}_0(x^{(i)})$.
Therefore, it is natural to choose $h_0^{(i)}$ as the hidden feature we condition on.

After controlling $H_0=h_0^{(i)}$, the meaning of the input node $X$ in the causal graph becomes all samples whose hidden feature is $h_0^{(i)}$.
Unfortunately, due to the sparsity in high dimensional spaces, only $x^{(i)}$ satisfies this constraint.
Intuitively, if we loosen this constraint a bit, the colliding effect will not disappear instantly.
Instead, the colliding effect will vanish gradually when the hidden feature becomes farther and farther away from $h_0^{(i)}$.
Put differently, colliding effects still exist when samples bear a resemblance to each other in the hidden space of the pre-trained model.

Now, we provide a derivation as follows:
\begin{small}
\begin{align}
   & \nonumber \textit{Effect}_{P}^{(i)} \\
   &= \mathbb{P}(\hat{Y}^{(i)}|H_0=h_0^{(i)},P=p) - \mathbb{P}(\hat{Y}^{(i)}|H_0=h_0^{(i)},P=0) \label{appendix:eq10} \\
   &= \sum_{k=1}^N ( \mathbb{P}(\hat{Y}^{(i)}|X=x^{(k)},H_0=h_0^{(i)}) \label{appendix:eq11}\\
   & \nonumber \qquad \qquad \mathbb{P}(X=x^{(k)}|H_0=h_0^{(i)},P=p) \\
   & \nonumber \qquad \qquad -\mathbb{P}(\hat{Y}^{(i)}|X=x^{(k)},H_0=h_0^{(i)})\\
   & \nonumber \qquad \qquad \mathbb{P}(X=x^{(k)}|H_0=h_0^{(i)},P=0) ) \\
   &= \sum_{k=1}^N \mathbb{P}(\hat{Y}^{(i)}|X=x^{(k)}) ( \mathbb{P}(X=x^{(k)}|H_0=h_0^{(i)},P=p) \label{appendix:eq12} \\
   & \nonumber \qquad \qquad - \mathbb{P}(X=x^{(k)}|H_0=h_0^{(i)},P=0) ) \\
   &\approx \sum_{k=1}^N \mathbb{P}(\hat{Y}^{(i)}|X=x^{(k)}) \mathbb{P}(X=x^{(k)}|H_0=h_0^{(i)},P=p) \label{appendix:eq13} \\
   &= \sum_{k=1}^N \mathbb{P}(\hat{Y}^{(i)}|X=x^{(k)}) \label{appendix:eq14} \\
   & \nonumber \frac{\mathbb{P}(H_0=h_0^{(i)}|X=x^{(k)},P=p)\mathbb{P}(X=x^{(k)}|P=p)}{\mathbb{P}(H_0=h_0^{(i)}|P=p)} \\
   &= \sum_{k=1}^N \mathbb{P}(\hat{Y}^{(i)}|X=x^{(k)}) W_P (x^{(i)}, x^{(k)}) \label{appendix:eq15} \\
   &\approx \sum_{k=0}^K \mathbb{P}(\hat{Y}^{(i)}|X=x^{(i,k)}) W_P (x^{(i)}, x^{(i,k)}) \label{appendix:eq16}
\end{align}
\end{small}
Eq.(\ref{appendix:eq10}) is deduced from Eq.(\ref{eq2}) and the condition of $H_0=h_0^{(i)}$. 
Eq.(\ref{appendix:eq11}) expands Eq.(\ref{appendix:eq10}) as the sum of all $N$ samples.
In Eq.(\ref{appendix:eq12}), $\mathbb{P}(\hat{Y}^{(i)}|X,H_0)=\mathbb{P}(\hat{Y}^{(i)}|X)$ because $X$ is the only mediator \citep{pearl2009causality} from $P$ to $\hat{Y}^{(i)}$.
In Eq.(\ref{appendix:eq13}), we approximate $\mathbb{P}(X=x^{(k)}|H_0=h_0^{(i)},P=0)$ as zero because the likelihood of $X=x^{(k)}$ is small when the model is randomly initialized.
Eq.(\ref{appendix:eq14}) is obtained by applying Bayes formula to Eq.(\ref{appendix:eq13}). 
In Eq.(\ref{appendix:eq14}), $\mathbb{P}(H_0=h_0^{(i)}|P=p)$ and $\mathbb{P}(X=x^{(k)}|P=p)$ are intractable and can be seen as constants.
We note that the likelihood term $\mathbb{P}(H_0=h_0^{(i)}|X=x^{(k)},P=p)$ represents how likely the hidden feature is $h_0^{(i)}$ when the input sample is $x^{(k)}$.
Obviously, the likelihood is the largest when $k=i$ and becomes smaller when the hidden feature of $x^{(k)}$ become farther away from $h_0^{(i)}$.
Therefore, the fractional term of Eq. \ref{appendix:eq14} can be regarded as a \textbf{scaling factor} of the likelihood $\mathbb{P}(\hat{Y}^{(i)}|X=x^{(k)})$.
In Eq.(\ref{appendix:eq15}), we re-write the fractional term of Eq.(\ref{appendix:eq14}) as a function of $x^{(i)}$ and $x^{(k)}$ since $h_0^{(i)}=\mathcal{F}_0(x^{(i)})$.
In Eq.(\ref{appendix:eq15}), we truncate the top K samples, which are closest to $x^{(i)}$, in the hidden space of the pre-trained model.
Besides, we let $x^{(i,0)}=x^{(i)}$ since $x^{(i)}$ has the largest similarity with itself.
Additionally, we let $\sum_{k=0}^K W_P(x^{(i)},x^{(i,k)}) = 1$ to ensure that the joint prediction is a probability distribution over categories.

\begin{table*}[t]
  \small
  \centering
  \caption{The dataset statistics.}
  \resizebox{\linewidth}{!}{
    \begin{tabular}{lcccccc}
    \toprule
          & \textbf{Train} & \textbf{Dev} & \textbf{Test} & \textbf{Option Number} & \textbf{Question Length} & \textbf{Option Length} \\
    \midrule
    CommonsenseQA & 8.5k  & 1.2k  & 1.2k  & 5     & 13.4  & 1.5 \\
    OpenBookQA & 5.0k  & 0.5k  & 0.5k  & 4     & 10.7  & 2.9 \\
    ARC-Easy & 2.2k  & 0.6k  & 2.4k  & 4     & 19.4  & 3.7 \\
    ARC-Challenge & 1.1k  & 0.3k  & 1.2k  & 4     & 22.3  & 4.9 \\
    QASC  & 7.3k  & 0.8k  & 0.9k  & 8     & 8.1   & 1.6 \\
    PIQA  & 14k   & 1.8k  & 1.8k  & 2     & 7.1   & 19.4 \\
    SocialIQA & 31k   & 1.9k  & 1.9k  & 3     & 20.1  & 3.6 \\
    \bottomrule
    \end{tabular}%
    }
  \label{tab:dataset_statistic}%
\end{table*}%

\section{Training Details}
\label{sec:appendix_b}
The dataset statistics is in Table \ref{tab:dataset_statistic}.
All models are implemented based on Pytorch \cite{paszke2019pytorch} and Huggingface \cite{wolf2019huggingface}.
We use the default hyper-parameters of RoBERTa and T5 according to the Huggingface implementation.
Following \citet{yasunaga-etal-2021-qa,khashabi-etal-2020-unifiedqa}, we concatenate all options as input when the backbone is T5 and concatenate each option respectively when the backbone is RoBERTa.
We tuned the batch size in \{64, 128\}, the learning rate of the backbone model in \{5e-5, 2e-5, 1e-5\}.
Before fine-tuning RoBERTa, a randomly initialized fully connected (FC) layer is added on top of RoBERTa, and the learning rate of the FC layer is 1e-2.
We use RAdam \citep{liu2019variance} as the optimizer and use a constant learning rate scheduler.
The weight decay is 1e-2, and the maximum gradient norm is 1.0.
For each dataset, the training hyper-parameters are the same for all methods for a fair comparison.
We select the best model according to the performance on the dev set and report the test accuracy of the chosen model.
The experiments are run on GeForce RTX 3090 GPU.
Each experiment is repeated five times.
Since we do not introduce any learnable parameters except PLMs, the total number of parameters of our method is the same as PLMs (RoBERTa-large and T5-large have 355M and 770M parameters, respectively).

\section{Details of the Competitive Fine-tuning Methods}
\label{sec:appendix_c}
The details of the competitive fine-tuning methods are as follows. 
Note that we use recommended hyper-parameters in the paper or the source code for a fair comparison.
\begin{itemize}
    \item vanilla fine-tuning: fine-tuning has been proven to be a simple and effective method of adapting large PLMs to downstream tasks. 
    \item BSS \citep{chen2019catastrophic} \footnote{https://github.com/thuml/Batch-Spectral-Shrinkage}: BSS focuses on mitigating negative transfer by penalizing the small singular values of the feature matrix. We penalize the smallest singular value, and the weight of the regularization term is set as 1e-3 as recommended.
    \item ChildTune-F\&ChildTune-D \citep{xu2021raise} \footnote{https://github.com/alibaba/AliceMind/tree/main/\\ChildTuning}: ChildTune-F\&ChildTune-D update a subset of parameters (called child network) of large PLMs in the backward process. ChildTune-D utilizes the Fisher Information Matrix estimated by the pre-trained model to determine the child network. ChildTune-F uses Bernoulli distribution to determine the child network. 
    \item Mixout \footnote{https://github.com/bloodwass/mixout} \citep{lee2019mixout}: Mixout randomly mixes the parameters of the pre-trained and the fine-tuned model to regularize the fine-tuning process. In the experiments, the mixing probability $p$ is set as 0.9.
    \item NoisyTune \citep{wu-etal-2022-noisytune}: NoisyTune adds uniform noises to the parameter of the pre-trained model based on their standard deviations. The scaling factor $\lambda$, which controls the relative noise intensity, is set as 0.15.
    \item R3F \footnote{https://github.com/facebookresearch/fairseq/tree/main/\\examples/rxf} \citep{aghajanyan2020better}: R3F alleviates representational collapse by introducing parametric noise. R3F generates noise from either a normal or uniform distribution.
    \item RecAdam \footnote{https://github.com/Sanyuan-Chen/RecAdam} \citep{chen-etal-2020-recall}: RecAdam optimizes a multi-task objective and utilize an annealing coefficient to gradually shift the objective from pre-training to downstream tasks. 
    \item ReInit \citep{zhang2020revisiting}: \citet{zhang2020revisiting} verified that transferring the top pre-trained layers slows down learning and hurts performance. ReInit re-initializes the top layers of PLMs when adapting to new tasks. In our experiments, we re-initialize the top 3 transformer block.
\end{itemize}

\section{Related Works of Commonsense QA}
\label{sec:related_works_of_commonsense_QA}
Commonsense reasoning is a key pillar of human cognition and intelligence, but it is still a long-standing challenge for deep learning systems \citep{xu-etal-2021-fusing,wang2020visual,talmor2018commonsenseqa}.
Current question and answering (QA) systems rely on external sources such as knowledge graphs (e.g., ConceptNet) \citep{yasunaga-etal-2021-qa,feng-etal-2020-scalable,wang-etal-2020-connecting,lin-etal-2019-kagnet}, knowledge bases (e.g., Wiktionary) \citep{xu-etal-2021-fusing} and generative pre-trained language models (e.g., GPT3 \citep{brown2020language}) \citep{liu-etal-2022-generated,yang-etal-2020-generative,rajani-etal-2019-explain,liu2022rainier}, and achieve remarkable success.
Despite the remarkable success, collecting high-quality external knowledge is usually expensive, and noisy knowledge is easily introduced \citep{liu-etal-2022-generated}.
In this paper, we present a novel fine-tuning method that retains commonsense knowledge from PLMs since they are exposed to a colossal amount of data in pre-training and inherently knowledge bases \citep{petroni-etal-2019-language,alkhamissi2022review}.
Different from the existing commonsense QA models, our method does not rely on KGs or GNNs.
Moreover, our method can be a plug-in module to enhance the performance of commonsense QA models. 
We compared six commonsense QA methods in the experiments:
\begin{itemize}
    \item Relation Network \citep{santoro2017simple} utilizes a relational reasoning structure over the knowledge graph;
    \item KagNet \citep{lin-etal-2019-kagnet} aggregates information with graph convolutional networks and LSTMs, and a hierarchical path-based attention mechanism;
    \item RGCN \citep{schlichtkrull2018modeling} extends the graph convolutional network with relation-specific weights;
    \item MHGRN \citep{feng-etal-2020-scalable} utilizes both GNNs and path-based models for commonsense QA;
    \item QAGNN \citep{yasunaga-etal-2021-qa} models the QA context and the knowledge graph in a joint graph and extracts their representations through a GNN.
    \item SAFE \citep{jiang2022great} designs a simple MLP-based knowledge encoder that utilizes statistical relation paths as features.
\end{itemize}

\begin{table}[t]
  \small
  \centering
  \caption{The average accuracy (\%) of fine-tuning and our method when T5-large is used as the backbone model.}
   \begin{tabular}{c|cc}
    \toprule
    Methods & Fine-Tuning & CET(Ours) \\
    \midrule
    CSQA  & 76.33 (0.55) & \textbf{76.85 (0.30)} \\
    OBQA  & 68.04 (0.62) & \textbf{69.14 (0.35)} \\
    ARC-Easy & 70.96 (0.48) & \textbf{71.63 (0.34)} \\
    ARC-Challenge & 46.68 (0.53) & \textbf{48.55 (0.58)} \\
    QASC  & 60.69 (0.78) & \textbf{61.79 (0.81)} \\
    PIQA  & 78.96 (0.42) & \textbf{81.58 (0.55)} \\
    SIQA  & 78.25 (0.38) & \textbf{79.40 (0.44)} \\
    \bottomrule
    \end{tabular}%
  \label{tab:table8}%
\end{table}%

\begin{figure*}[t]
    \centering
    \subfloat[The backbone is RoBERTa-large]{
        \includegraphics[width=0.5\linewidth]{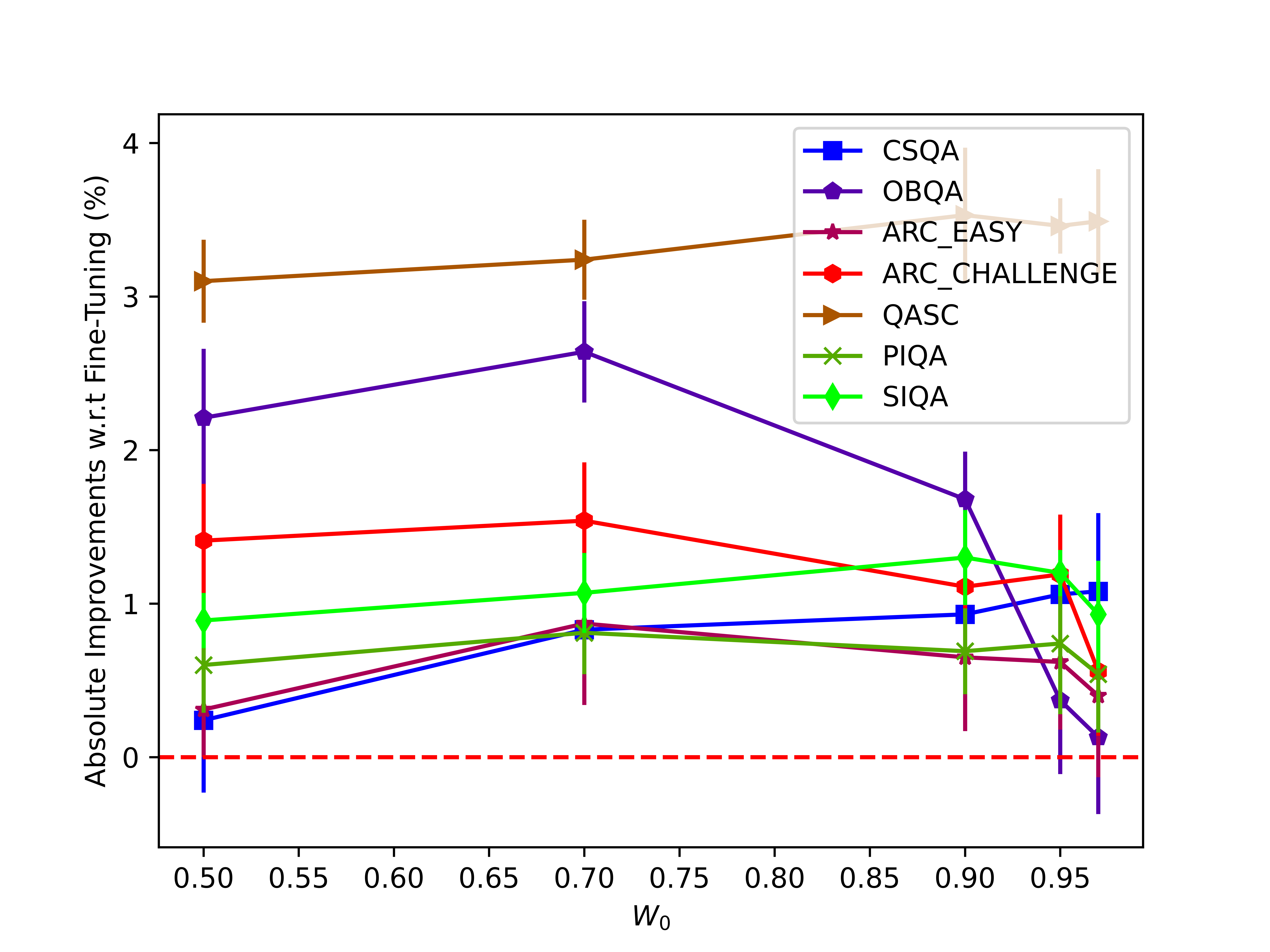}
        \label{fig3}
    }
     \subfloat[The backbone is T5-large]{
        \includegraphics[width=0.5\linewidth]{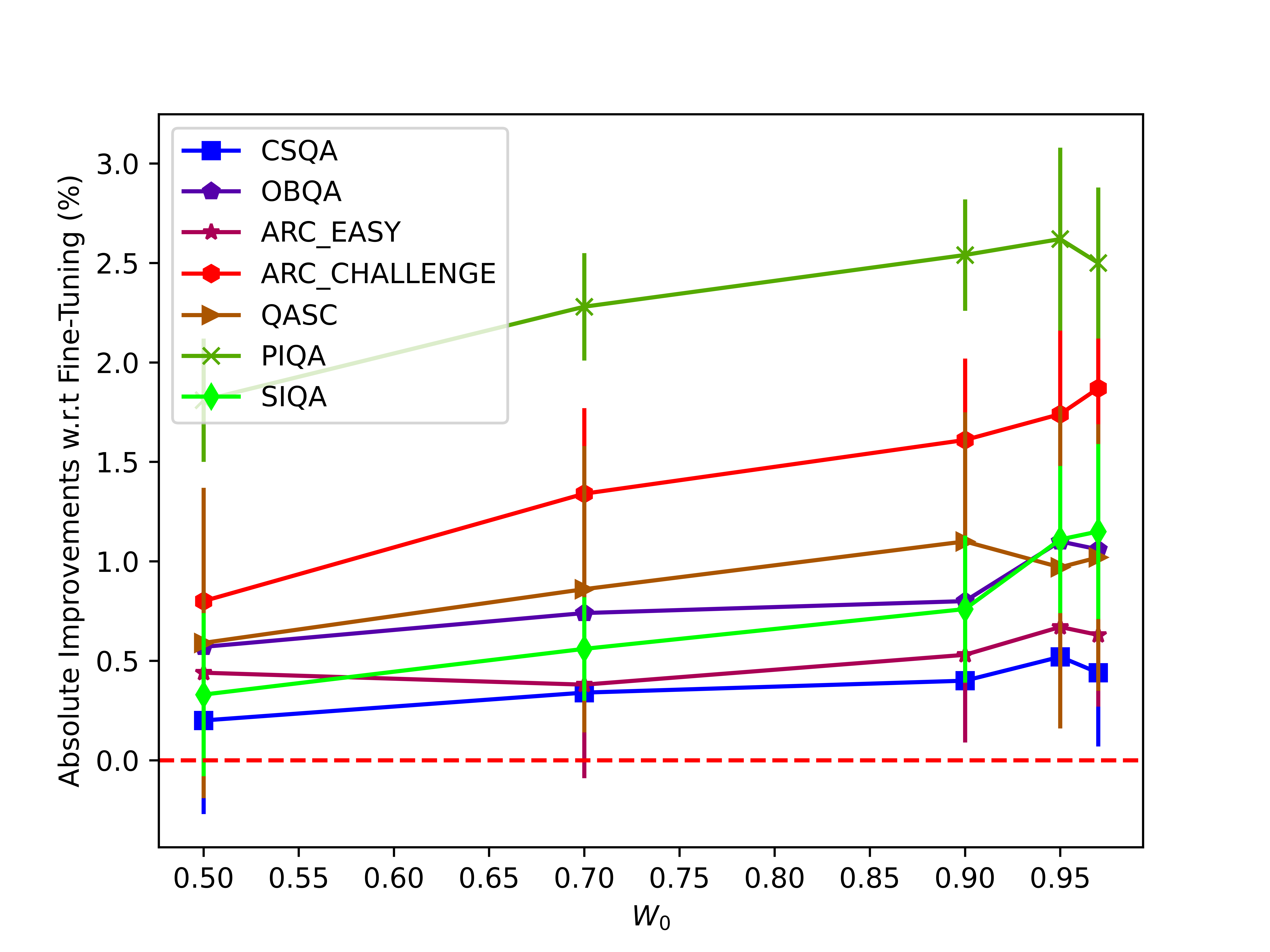}
        \label{fig4}
    }
    \caption{The absolute improvements (\%) of our method w.r.t fine-tuning when $W_0=\{0.50,0.70,0.90,0.95,0.97\}$. The backbone model is RoBERTa-large (a) and T5-large (b), respectively.}
\end{figure*}

\begin{table}[t]
  \small
  \centering
  \caption{The average accuracy (\%) of our method when different $K$ is selected.}
    \begin{tabular}{l|cc}
    \toprule
          & K=3   & K=5 \\
    \midrule
    CSQA  & 76.74  & \textbf{76.82 } \\
    OBQA  & \textbf{70.88 } & 70.76  \\
    ARC-EASY & \textbf{68.59 } & 68.53  \\
    ARC-CHALLENGE & 47.40  & \textbf{47.52 } \\
    QASC  & 57.42  & \textbf{57.57 } \\
    PIQA  & 79.13  & \textbf{79.43 } \\
    SIQA  & 78.61  & \textbf{78.76 } \\
    \bottomrule
    \end{tabular}%
  \label{tab:table3}%
\end{table}%

\section{Additional Experimental Results}
\label{sec:additional_experimental_results}
\noindent\textbf{Experiments on T5.}\quad
Our method is model-agnostic since it only requires computing the joint prediction.
Different from discriminant models such as RoBERTa, T5 is a generative model whose output is in text format.
Following \citet{khashabi-etal-2020-unifiedqa}, we concatenate a question and its all options with prefixes $(a),(b),(c),...$ as the input, and expect the model to output the ground-truth option in text format.
To adapt our model to T5, we substitute the prediction from the probability distribution over options to the probability distribution over vocabulary.
In this way, we encourage T5 to generate the same gold answer when the input is the question of the anchor sample and its KNNs.

The experimental result is in Table \ref{tab:table8}.
From the result, we find that our method still improves vanilla fine-tuning consistently, which demonstrates that our approach can be applied to various architectures.
Besides, we also apply ReInit on T5 as in RoBERTa.
Unfortunately, T5 fails to adapt to downstream tasks when only a few parameters are re-initiated (e.g., the self-attention layer or the cross-attention layer in the topmost transformer block).
We conjecture that the final language modeling head (LM head), which maps the last hidden states to the vocabulary space, hinders the knowledge of the bottom layers to transfer to new tasks.
Different from ReInit, our method is also applicable to T5 because it has no assumptions about the model architecture.

\noindent\textbf{Hyper-parameter Analysis.}\quad
We consider two hyper-parameters that may influence the effectiveness of our method: the number of neighbors $K$ and the weight for controlling the strength of colliding effects $W_0$.
Fig. \ref{fig3} and \ref{fig4} show that our method is robust when various $W_0$ are chosen.
When the backbone is RoBERTa-large, our method achieves the best performance when $W_0=0.7$ on OBQA, ARC-Easy, and ARC-Challenge; when $W_0=0.9$ on QASC and SIQA; and when $W_0=0.97$ on CSQA.
When the backbone is T5-large, our method achieves the best performance when $W_0=0.9$ on QASC; when $W_0=0.95$ on CSQA, OBQA, ARC-Easy, and PIQA; and when $W_0=0.97$ on ARC-Challenge and SIQA.
In addition, we find that some datasets, such as CSQA, require more domain-specific knowledge while some datasets, such as OBQA, require more pre-trained knowledge.
The result of $K$ in Table \ref{tab:table3} shows that a larger $K$ is beneficial.
Our method is also robust to $K$ because the similarity threshold $\theta$ truncates the number of nearest neighbors for each sample.

\begin{table*}[!t]
  \centering
  \caption{Comparison between CET and vanilla fine-tuning on NER.}
     \resizebox{0.8\linewidth}{!}{
    \begin{tabular}{lcccccc}
    \toprule
    \multicolumn{1}{c}{\multirow{2}[4]{*}{\textbf{Method}}} & \multicolumn{2}{c}{\textbf{CoNLL2003}} & \multicolumn{2}{c}{\textbf{OntoNotes5}} & \multicolumn{2}{c}{\textbf{I2B2}} \\
\cmidrule{2-7}          & \textbf{Micro F1} & \textbf{Macro F1} & \textbf{Micro F1} & \textbf{Macro F1} & \textbf{Micro F1} & \textbf{Macro F1} \\
    \midrule
    Vanilla Fine-Tuning & 92.52 & 91.09 & 89.35 & 80.42 & 92.81 & 85.61 \\
    CET (Ours) & \textbf{92.94} & \textbf{91.52} & \textbf{90.09} & \textbf{81.67} & \textbf{94.07} & \textbf{88.46} \\
    \bottomrule
    \end{tabular}%
    }
  \label{tab:exp_ner}%
\end{table*}%

\noindent\textbf{Differences between Our Method and Data Augmentation.}\quad
Our method recombines the KNN questions with the options of the anchor sample.
A reasonable conjecture is that our method ``adds'' KNN samples to enhance generalization ability.
We do the following experiment to test the hypothesis: We add the same KNN samples generated by our method into the original training set for fine-tuning.
The result shows that its improvement is not statistically significant.
The reason may be as follows:
Recall that we set $\theta=1.0$ on five out of six datasets where the gold answer of the anchor sample and its KNNs should be matched precisely.
Therefore, on most datasets, the KNN samples recombine with the options containing their original gold answer, suggesting that they provide no additional information.
Besides, the newly added samples change the data distribution of the original training set.

\noindent\textbf{Experiments on Named Entity Recognition.}\quad
To demonstrate that CET has the potential to improve more generic tasks, we apply CET to another task, Named Entity Recognition (NER), which is a fundamental task in NLP.
First, NER can be formulated as a word-level classification task. 
Therefore, both "anchor" and KNNs refer to a specific word. 
Then, we use the Euclidean distance as a metric to find the KNNs in the space of the last hidden states of PLMs. 
Considering NER focuses on recognizing entities, we only compute the causal effects on entity words. 
During training, both the sentences containing anchor and KNN words are fed into the model. 
And then, we compute the joint prediction as in Eq.\ref{eq6} by combining the score prediction of the anchor word and the corresponding KNN words. 
Finally, we jointly optimize the causal effects of entity words and the vanilla fine-tuning objective of non-entity words as in Eq.\ref{eq9}. 

We choose three widely used datasets for experiments: CoNLL2003 \cite{sang2003introduction}, Ontonotes5 \cite{hovy2006ontonotes}, I2B2 \cite{murphy2010serving}.
Following previous experiments, we use RoBERTa-large as the backbone.
The result in Table \ref{tab:exp_ner} indicates that CET outperforms vanilla fine-tuning consistently.

To better understand CET, here is an example from CoNLL2003:
The anchor is a Location entity "California" in the sentence "…Marine Laboratories in California say …".
Its three nearest neighbours are
1. "California" in the sentence "At California, Tim …";
2. "Oakland" in the sentence "OAKLAND AT NEW YORK";
3. "Florida" in the sentence "At Florida, …". 
As shown, the anchor and KNN words share the related prior knowledge of PLMs, which can also be illustrated in Figure \ref{fig2}.

\section{More examples of Colliding Effects}
\label{sec:more_examples_of_colliding_effects}

\begin{table}[htbp]
  \small
  \centering
  \caption{Examples from PIQA and QASC.}
    \resizebox{0.95\linewidth}{!}{
    \begin{tabular}{ccl}
    \toprule
    PIQA  & Gold Answer & \multicolumn{1}{c}{Question} \\
    \midrule
    Anchor & throw it away & how do you dispose of a cutip? \\
    \midrule
    \multirow{2}[4]{*}{KNNs} & throw it away & how do you dispose of something? \\
\cmidrule{2-3}          & throw it away & how do you scrap metal? \\
    \bottomrule
    \end{tabular}%
    }
    \\
    \resizebox{0.95\linewidth}{!}{
    \begin{tabular}{ccl}
    \toprule
    QASC  & Gold Answer & \multicolumn{1}{c}{Question} \\
    \midrule
    Anchor & bacteria & What causes botulism? \\
    \midrule
    \multirow{5}[10]{*}{KNNs} & bacteria & what may die if it becomes too hot? \\
\cmidrule{2-3}          & bacteria & what causes serious illness? \\
\cmidrule{2-3}          & bacteria & What causes food to spoil? \\
\cmidrule{2-3}          & bacteria & What can cause people to die? \\
\cmidrule{2-3}          & bacteria & what feed on dead organisms? \\
    \bottomrule
    \end{tabular}%
    }
  \label{tab:table6}%
\end{table}%

\begin{table*}[t]
  \small
  \centering
  \caption{Examples from CSQA, OBQA, ARC-Easy, ARC-Challenge, and SIQA.}
    \begin{tabular}{ccl}
    \toprule
    CSQA & Gold Answer & \multicolumn{1}{c}{Question} \\
    \midrule
    Anchor & television & To prevent any glare during the big football game he made sure to clean the dust of his what? \\
    \midrule
    \multirow{5}[9]{*}{KNNs} & television & Where do you watch garbage? \\
\cmidrule{2-3}          & television & What home entertainment equipment requires cable? \\
\cmidrule{2-3}          & television & What does one watch garbage reality shows on? \\
\cmidrule{2-3}          & television & Where might I hear and see information on current events? \\
\cmidrule{2-3}          & television & James wasn't a repair person, but even he knew that he didn't need a  freon coin in a what? \\
    \bottomrule
    \end{tabular}%
    \\
    \begin{tabular}{ccl}
    \toprule
    OBQA  & Gold Answer & \multicolumn{1}{c}{Question} \\
    \midrule
    Anchor & sun   & The leaves of a plant benefit from? \qquad \qquad \qquad \qquad \qquad \qquad \qquad \qquad \qquad \qquad \qquad \qquad \quad\\
    \midrule
    \multirow{4}[8]{*}{KNNs} & sun   & The moon orbits an object that orbits the \\
\cmidrule{2-3}          & sun   & Which of these items is required for a deer to live \\
\cmidrule{2-3}          & sun   & What is larger then the human planet and causes cycles of day and night? \\
\cmidrule{2-3}          & the sun & Despite what some think, instead around themselves, our planet spins around \\
    \bottomrule
    \end{tabular}%
    \\
    \begin{tabular}{ccl}
    \toprule
    ARC-Easy & Gold Answer & \multicolumn{1}{c}{Question} \\
    \midrule
    Anchor & line graph & \tabincell{l}{A student wants to find the relationship between the diameter of several plastic disks \\ and the circumference of each disk. Which of these types of graphs should be constructed \\ to determine this relationship?} \\
    \midrule
    \multirow{5}[10]{*}{KNNs} & line graph & \tabincell{l}{The number of squirrels in a certain ecosystem changes over time. These changes can be \\ represented as a number of connected data points. Which method would a student most \\ likely use to show this information?} \\
\cmidrule{2-3}          & line graph & \tabincell{l}{In a city, the daily high and low 16 temperatures for a month are best represented by which \\ of the following?} \\
\cmidrule{2-3}          & line graph & \tabincell{l}{A student measures the growth of a group of plants given different amounts of fertilizer. \\ Which data display should the student use to compare the growth of the plants?} \\
\cmidrule{2-3}          & line graph & \tabincell{l}{Scientists recorded the hourly temperature at a weather station for the month of July and \\ want to quickly measure a trend over time in temperature changes. Which of these formats \\ would be the most appropriate representation of the temperature data to quickly measure \\ any trend?} \\
\cmidrule{2-3}          & line graph & \tabincell{l}{The most effective way to show a change happening over time is to display your results \\ using a} \\
    \bottomrule
    \end{tabular}%
    \\
    \begin{tabular}{ccl}
    \toprule
    ARC-Challenge & Gold Answer & \multicolumn{1}{c}{Question} \\
    \midrule
    Anchor & air   & \tabincell{l}{Four materials are put into small containers. These materials are then moved from the \\ small containers into larger containers. Which material will spread out to completely \\ fill a larger container?} \\
    \midrule
    \multirow{4}[8]{*}{KNNs} & air   & When you make soap bubbles, what is inside the bubbles? \\
\cmidrule{2-3}          & air   & When a tadpole grows, its gills change into lungs. What does it now need to survive? \\
\cmidrule{2-3}          & air   & How are green plants an important part of the carbon dioxide-oxygen cycle? \\
\cmidrule{2-3}          & air   & Which of the following substances can be separated into several elements? \\
    \bottomrule
    \end{tabular}%
    \\
    \begin{tabular}{ccl}
    \toprule
    SIQA  & Gold Answer & \multicolumn{1}{c}{Question} \\
    \midrule
    Anchor & compassionate & \tabincell{l}{Jan had always wanted a puppy, but decided to adopt an older shelter dog instead. How would \\ you describe Jan?} \\
    \midrule
    \multirow{5}[10]{*}{KNNs} & compassionate & \tabincell{l}{Jan gave Kai's husband a hug after hearing the good news about Kai's recovery. How would \\ Kai feel as a result?} \\
\cmidrule{2-3}          & compassionate & Quinn ran over a squirrel on the road. They felt a little guilty. How would you describe Quinn? \\
\cmidrule{2-3}          & compassionate & \tabincell{l}{Cameron was volunteering at a soup kitchen and provided assistance to individuals. How would \\ Cameron feel afterwards?} \\
\cmidrule{2-3}          & compassionate & \tabincell{l}{Bailey found out that the local fire department lacked funding. Bailey decided to do something \\ about it. How would you describe Bailey?} \\
\cmidrule{2-3}          & compassionate & \tabincell{l}{Ash let the dog inside as it was getting too hot for dog to be outside. How would you describe \\ Ash?} \\
    \bottomrule
    \end{tabular}%
  \label{tab:table5}%
\end{table*}%

\end{document}